\pdfoutput=1

\documentclass[11pt]{article}

\usepackage[final]{acl}

\usepackage{times}
\usepackage{latexsym}
\usepackage{bm}

\usepackage[T1]{fontenc}

\usepackage[utf8]{inputenc}

\usepackage{microtype}

\usepackage{inconsolata}
\usepackage{graphicx}
\usepackage{booktabs}
\usepackage{bm}
\usepackage{colortbl}
\usepackage{balance}
\usepackage{booktabs} 
\usepackage{graphics}
\usepackage{array}
\usepackage{flushend}
\usepackage[english]{babel}
\usepackage[T1]{fontenc}
\usepackage{textcomp}
\usepackage{latexsym}
\usepackage{natbib}
\usepackage{graphicx}
\usepackage{multirow}
\usepackage[labelformat=simple]{subcaption}
\usepackage{algorithm}
\usepackage{algpseudocode}
\usepackage{amsmath}
\usepackage{tabularx}
\usepackage{pifont}
\usepackage{amsfonts}
\usepackage{enumitem}
\usepackage{pifont}
\usepackage{bbding}

\title{Enabling Discriminative Reasoning in LLMs for Legal Judgment Prediction}

\author{Chenlong Deng$^{1}$, Kelong Mao$^{1}$, Yuyao Zhang$^{1}$, Zhicheng Dou$^{1}$\thanks{Corresponding author.}\\ 
    $^1$Gaoling School of Artificial Intelligence, Renmin University of China \\ 
    \texttt{\{dengchenlong,dou\}@ruc.edu.cn} \\
}

\begin{document}
\maketitle
\begin{abstract}
Legal judgment prediction is essential for enhancing judicial efficiency. In this work, we identify that existing large language models (LLMs) underperform in this domain due to challenges in understanding case complexities and distinguishing between similar charges. 
To adapt LLMs for effective legal judgment prediction, we introduce the \textit{\textbf{A}sk-\textbf{D}iscrimin\textbf{A}te-\textbf{P}redic\textbf{T}} (ADAPT) reasoning framework inspired by human judicial reasoning.
ADAPT involves decomposing case facts, discriminating among potential charges, and predicting the final judgment. 
We further enhance LLMs through fine-tuning with multi-task synthetic trajectories to improve legal judgment prediction accuracy and efficiency under our ADAPT framework. 
Extensive experiments conducted on two widely-used datasets demonstrate the superior performance of our framework in legal judgment prediction, particularly when dealing with complex and confusing charges.
\end{abstract}

\section{Introduction}
\begin{figure*}[!t]
	\centering
	\includegraphics[width=\linewidth]{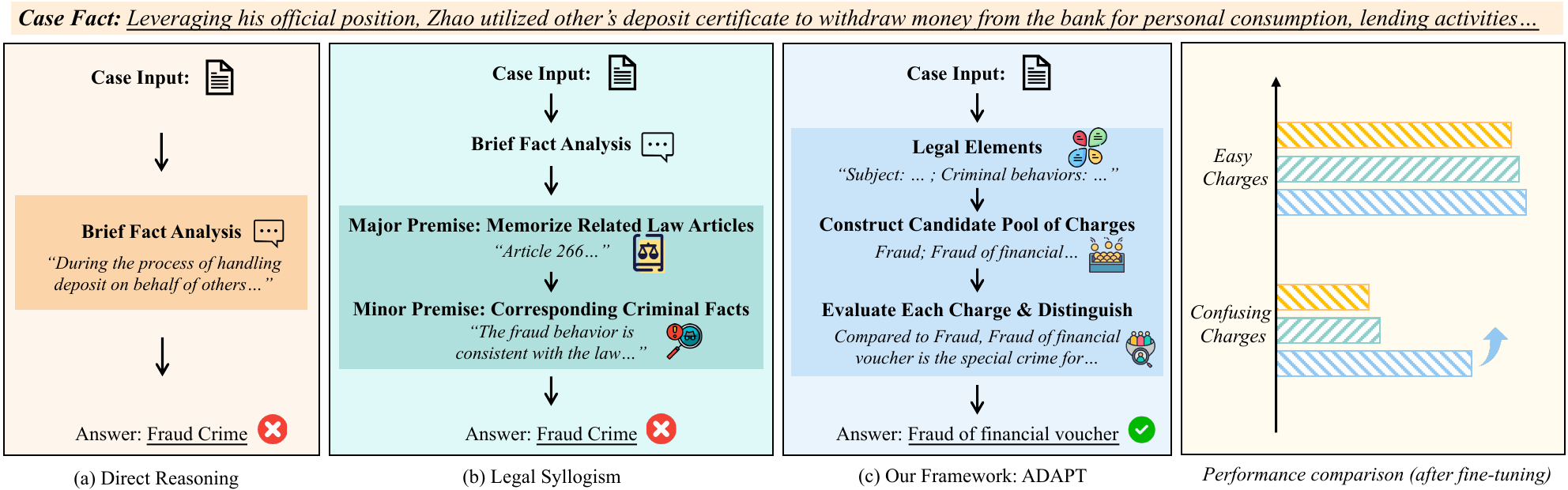}
	\caption{Comparison of our framework with direct reasoning and legal syllogism. We notice that our approach improves the performance on confusing charges more obviously after fine-tuning.}
	\label{fig: introduction}
 \vspace{-2ex}
\end{figure*}

Legal judgment prediction (LJP) is a key research area within the legal natural language processing (NLP) community, aiming to provide automated reference judgments to help judges and other professionals manage cases more efficiently~\cite{DBLP:conf/emnlp/LuoFXZZ17, DBLP:conf/emnlp/ChalkidisFMAA20, DBLP:conf/acl-nllp/NiklausCS21}. The main challenges in enhancing judgment prediction systems are twofold: understanding case facts and distinguishing between similar charges. Understanding case facts involves extracting key information from complex descriptions~\cite{DBLP:conf/sigir/Yue0JWZACYW21}, while distinguishing between charges requires identifying the correct labels among confusing options~\cite{DBLP:conf/acl/XuWCPWZ20}. To tackle these issues, researchers have explored the use of advanced language models, the incorporation of external legal knowledge, and the reference to precedents to enhance model performance~\cite{DBLP:conf/ijcai/ZhaoGXZC22}.

Recently, large language models (LLMs) have achieved state-of-the-art performance across a range of tasks due to their expanded parameter scales and training data~\cite{DBLP:journals/corr/abs-2303-18223, DBLP:journals/corr/abs-2308-07107, DBLP:journals/fcsc/WangMFZYZCTCLZWW24, DBLP:journals/corr/abs-2301-08745}. These models also display emergent abilities, such as instruction following and in-context learning, allowing them to quickly adapt to specific tasks based on minimal instructions or examples~\cite{DBLP:conf/nips/BrownMRSKDNSSAA20, DBLP:journals/corr/abs-2303-08774}. Although preliminary evaluations of mainstream LLMs in legal judgment prediction have been conducted, results indicate that their performance still lags behind many traditional supervised methods, suggesting that the existing LLMs are still far from being good judgment predictors~\cite{DBLP:conf/emnlp/Shui00C23, DBLP:conf/emnlp/VatsZDSBNGRG23, DBLP:conf/icail/JiangY23}.

Through our experiments, we find that current LLMs still struggle significantly more with distinguishing ``confusing charges'', which refer to charges that have similar or even overlapping key behaviors with other charges.  This highlights a critical aspect of the LJP task: \textit{certain criminal behaviors can satisfy parts or even all of the conditions for multiple charges, creating ambiguity}. Due to the lack of extensive domain knowledge and reasoning training in the legal context, existing LLMs exhibit insufficient reasoning capability to effectively differentiate between similar charges.


To adapt LLMs for effective legal judgment prediction, we introduce the \textit{\textbf{A}sk-\textbf{D}iscrimin\textbf{A}te-\textbf{P}redic\textbf{T}} (ADAPT) reasoning framework in this paper. 
Our framework is inspired by the thought process of human judges, who use legal knowledge to navigate between facts and norms, as described by the classic phrase: \textit{``the gaze shuttles back and forth between facts and norms''}~\cite{ruthers2013rechtstheorie}.
Specifically, In the first step, \textit{Ask}, we decompose the noisy case description into multiple aspects under legal theory to clarify the key criminal facts. In the second step, \textit{Discriminate}, the model uses its parameterized knowledge to generate a candidate pool of the most probable charges and relevant law articles. Within this pool, the model further differentiates among the candidates, assessing the degree of alignment between each candidate and the criminal facts. Finally, in the \textit{Predict} step, the model synthesizes the previous reasoning process to provide the final prediction.

Experimental results show that our ADAPT prompting framework outperforms traditional prompting methods, such as direct reasoning and legal syllogism (Figure~\ref{fig: introduction}), in leveraging LLMs for legal judgment prediction. However, LLMs still struggle to consistently generate accurate ADAPT reasoning trajectories so as to finally make correct predictions. We hypothesis that this limitation is due to their narrow legal-specific knowledge and lack of familiarity with our specialized ADAPT reasoning patterns. Furthermore, we find that current LLMs frequently avoid providing sentencing ranges because of strict safety alignment protocols, which hinders their ability to complete this essential task in legal judgment prediction.

To address these issues, we further propose fine-tuning an enhanced LLM within our ADAPT framework for more comprehensive, efficient, and effective legal judgment prediction. Specifically, we strengthen the LLM by incorporating additional context labels—such as discriminative labels, charges, legal articles, and sentencing ranges—and prompt it to generate high-quality synthetic reasoning trajectories tailored to our ADAPT famework. We then use these multi-task synthetic trajectories to fine-tune a smaller LLM, enabling it to perform accurate reasoning under our ADAPT framework.

We conduct extensive experiments on two datasets, CAIL2018 and MultiLJP, which belong to single-defendant and multi-defendant scenarios, respectively. The results show that our approach achieves new state-of-the-art in both scenarios, especially on the most challenging set of charges.

Our contributions are summarized as:

(1) We pinpoint that the underperformance of LLMs in legal judgment prediction primarily stems from their difficulty in distinguishing between confusing charges. 

(2) We propose the ADAPT reasoning framework to emulate human judicial reasoning, which guides LLM to navigate between legal facts and norms to improve the accuracy of legal judgments.

(3) We fine-tune an enhanced LLM using knowledge distillation on multi-task synthetic trajectories to achieve more comprehensive, efficient, and effective legal judgment prediction under our ADAPT framework.


\begin{figure*}[!t]
	\centering
	\includegraphics[width=\linewidth]{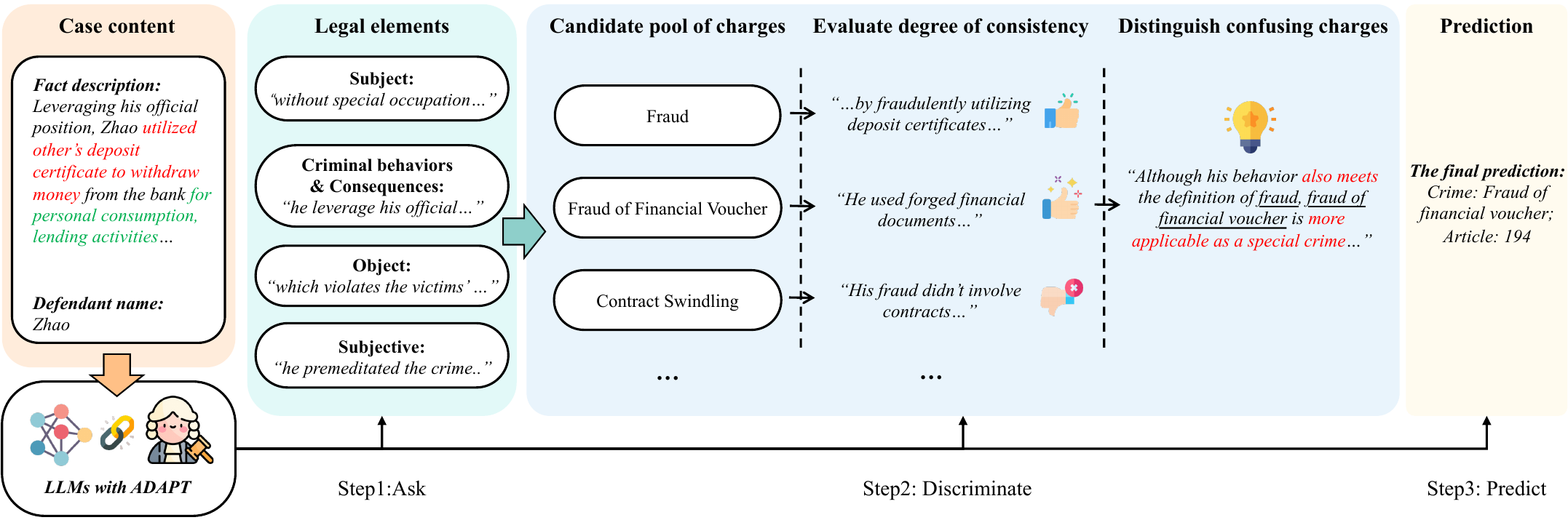}
	\caption{Overview of our framework. The final judgment is predicted based on three different reasoning steps.}
	\label{fig: method}
 \vspace{-2ex}
\end{figure*}

\section{Related Work}

\paragraph{Legal judgment prediction}
Legal judgment prediction is a long-standing legal NLP task. The evolution of this task's technology has transitioned through various phases: initially relying on rule-based approaches~\cite{nagel1963applying, segal1984predicting}, advancing to statistical machine learning techniques~\cite{katz2017general, DBLP:conf/icail/SuleaZMVDG17}, and currently dominated by deep learning methodologies~\cite{DBLP:conf/acl/XuWCPWZ20, DBLP:conf/sigir/Yue0JWZACYW21, DBLP:journals/tois/ZhangDZW23}. Additionally, incorporating domain-specific legal knowledge~\cite{DBLP:conf/ijcai/ZhaoGXZC22} or precedents~\cite{DBLP:conf/cncl/ZhangD23, DBLP:conf/emnlp/WuZL0LZS0K23} is also an important direction in existing research. 
With the continuous advancement of methods, this task has expanded from a simplified multi-class classification problem to complex scenarios that mirror real-world situations, such as dealing with multiple defendants~\cite{DBLP:conf/emnlp/LyuH0ZGRCWR23} and multiple law articles~\cite{DBLP:conf/sigir/LiuWZS0WK23}. In this context, we explore the use of large language models as the foundation model and conduct robust reasoning under the setting of multi-label classification.

\paragraph{Reasoning skills in language models}
Recent studies have shown that effective reasoning can be achieved in LLMs by using prompting techniques such as ``Chain-of-Thought''~\cite{DBLP:conf/nips/Wei0SBIXCLZ22} and ``Self-Ask''~\cite{DBLP:conf/emnlp/PressZMSSL23}. In the legal context, previous study shows that legal syllogism can enhance the performance of LJP~\cite{DBLP:conf/icail/JiangY23}. Furthermore, some research attempts focus on distilling the reasoning processes of large models into smaller models, thereby achieving approximate reasoning capabilities at a lower cost~\cite{DBLP:conf/acl/HoSY23, DBLP:journals/corr/abs-2306-02707}. Our approach synthesizes the trajectory of the ADAPT framework to fine-tune a 7B model, enabling it to achieve both robust and efficient reasoning.

\section{Methodology}

Existing LLMs mostly rely on direct answering or judicial syllogism for reasoning, which requires the model to directly provide the correct law articles and charges. This exceeds the capabilities of these language models and even human experts, leading to bad performance.
In this work, we enable LLM to emulate the reasoning pattern of real-world judges to conduct discriminative reasoning for accurate legal judgment prediction.

\subsection{Preliminaries}
We first formally describe the workflow of legal judgment prediction. Given the criminal fact $f$ from a real case, the name of one of the defendants, a set of charges $\mathcal{C}$, and a set of law articles $\mathcal{A}$, our task is to predict the applicable subset of charges $\mathcal{C}_d$, the relevant subset of law articles $\mathcal{A}_d$, and the term of imprisonment for the defendant $d$. To align with legal practice, we follow recent studies that treat charge and law article prediction as multi-label classification tasks, and term of imprisonment prediction as a multi-class classification task.

\subsection{Discriminative Reasoning Framework}
We propose a discriminative reasoning framework, called ADAPT, to guide LLM to gradually deduce the most appropriate charges and law articles step by step, including \textit{Ask}, \textit{Discriminate}, and \textit{Predict}.
As shown in Figure~\ref{fig: method}, in the first step, \textit{Ask}, the model is prompted to identify the key legal elements that constitute crimes through a question-answering approach. Then,  in the second step, \textit{Discriminate}, we prompt LLM to utilize the extracted key elements from the \textit{Ask} step to initially identify the top-$K$ most possible charges, and subsequently evaluate the consistency between each charge and the established facts. Finally, in the \textit{Predict} step, LLM integrates the reasoning signals from the previous two steps to identify the most suitable charges and law articles. 
We describe the details of these three steps in the following.

\paragraph{Step1: Ask} The objective of the \textit{Ask} step is to clarify the key elements that constitute crimes. We use legal theory~\cite{ruthers2013rechtstheorie} to guide LLM to summarize four aspects from the facts: (1) \textit{Subject}, which refers to the defendant's occupation and identity characteristics, such as state officials. (2) \textit{Criminal behaviors and consequences}, which contains the defendant's specific actions and resulting harm. (3) \textit{Object}, which is the entities or legal interests violated by the criminal acts. (4) \textit{Subjective aspect}, which is the psychological state of the defendant, such as direct purpose, negligence, and so on. Inspired by \cite{zhang2023aligning}, we prompt in question-answering form for accurate summarization. 

\paragraph{Step2: Discriminate} To avoid invalid reasoning caused by selecting from incorrect confusing charges, it is necessary to carefully distinguish candidate charges before making predictions. Specifically, we prompt the LLM to first provide several most likely candidate charges based on its parametric knowledge. Based on these candidates, the model then evaluates the consistency of each candidate with the key elements and distinguishes the main differences between these charges.

\paragraph{Step3: Predict} By contextualizing the reasoning trajectories of the initial two steps, the LLM is prompted to predict the final judgment result.

\subsection{Improving ADAPT with Fine-tuning}
While our ADAPT prompting framework outperforms traditional prompting methods such as direct reasoning or legal syllogism, we find that current general LLMs still struggle to consistently generate fully accurate reasoning trajectories. This limitation arises from their restricted legal-specific knowledge and lack of exposure to our specialized ADAPT reasoning patterns.
Additionally, we find that current LLMs often reject giving the sentencing ranges, likely because of their strict safety alignment. 
To address these issues, we propose fine-tuning a better LLM under our ADAPT framework for more comprehensive, efficient, and effective legal judgment prediction.

\paragraph{Synthetic trajectories generation}
we first generate synthetic ground-truth reasoning trajectories for the three steps of ADAPT using a larger model, specifically a 72B parameter LLM, which is provided with refined instructions and additional context labels, such as ground-truth discriminative labels, charges, and relevant legal articles. By incorporating these additional context labels, we find that the 72B LLM is able to generate highly accurate reasoning trajectories for each step.
Subsequently, we use these high-quality synthetic reasoning trajectories to fine-tune a smaller LLM.

\paragraph{Multi-task instruction tuning}
We have five specific tasks in our fine-tuning uniformly using the language modeling loss function:
\begin{eqnarray*}
    \mathcal{L}_{\text{task}} = -\frac{1}{T_{\text{task}}} \sum_{t=1}^{T_{\text{task}}} \log P_{\theta}(y^{\text{task}}_{t}|\mathcal{F}_{\text{task}}(x_{\text{task}}), y^{\text{task}}_{\textless t}),
\end{eqnarray*}
where $\mathcal{F}$ is the task-specific prompting function for formulating the input instruction.
$x$, $y$, and $T$ are the input, target response, and the number of tokens in the response, respectively.

Specifically, the first two tasks are \textit{ask} and \textit{discriminate}, corresponding to the first two steps of ADAPT.
The input $x_{\text{ask}}$ consists of the criminal fact $f$ and the specified defendant $d$. The input $x_{\text{disc}}$ additionally contains 
$y^\text{ask}$, which is the target output of the \textit{Ask} step generated by the 72B LLM.

The third task is \textit{sentencing}, which is to improve the model's perception of the sentencing factors. 
Its input $x_{\text{sent}}$ consists of a set of charges $C_d$ against the defendant $d$;
The fourth task is \textit{article}, which is to improve the model comprehension of the correspondence between the case facts and the specified law articles. The model learns to recite the content of the given article numbers as well as explain in detail how the defendant's actions align with these articles. Its input $x_\text{article}$ contains the criminal fact $f$, the specified defendant $d$, and the article number.
The training targets of these two tasks are also generated by employing the 72B LLM with additional context labels, including the articles and the sentencing ranges.

Finally, the last task, \textit{predict\_all}, is to contextualize all of the previous reasoning results and predict the final charges, articles, and sentencing ranges just in one prompt.
Its input $x_\text{predict\_all}$ contains the criminal fact $f$ and the specified defendant $d$.
Its training target $y_{\text{predict\_all}}$ is the concatenation of the synthetic reasoning trajectories of the three steps of ADAPT and the sentence range labels.

For clarity, we show the task-specific prompting functions and all synthetic prompts of different tasks in Appendix~\ref{Appendix: Prompts}.
We equally mix the training samples of different tasks to perform multi-task fine-tuning.

\section{Experiments}

\subsection{Datasets and Evaluation Metrics}
We conduct extensive experiments in both single-defendant and multi-defendant scenarios to comprehensively evaluate our method's performance in real-world applications. In the single-defendant context, we employ the widely-used CAIL2018 dataset~\cite{DBLP:journals/corr/abs-1807-02478}.
For the multi-defendant case, we select the MultiLJP~\cite{DBLP:conf/emnlp/LyuH0ZGRCWR23} dataset whose labels are verified by human experts. Both datasets are divided into 11 intervals to convert prison term prediction to a multi-class classification task. Detailed statistics of both datasets are provided in Table~\ref{table: dataset}. For evaluation metrics, we follow previous works to adopt Accuracy (Acc.), Macro Precision (Ma-P), Macro Recall (Ma-R), and Macro F1 (Ma-F) across all sub-tasks.

\begin{table}[!t]    
    \scalebox{0.82}{\begin{tabular}{p{0.3\textwidth}p{0.1\textwidth}p{0.1\textwidth}}
        \toprule
        Dataset & CAIL2018 & MultiLJP \\
        \midrule
        \# Train cases & 118,399 & 18,968\\
        \# Test cases & 1,120 & 2,370\\
        \# Charges & 191 & 23\\
        \# Articles & 162 & 22\\
        \# Intervals of prison term & 11 & 11 \\
        \# Average criminal per case & 1 & 3.71 \\
        Average length per case & 440.9 & 3,040.8\\
        \bottomrule
    \end{tabular}}
\caption{Basic statistics of the datasets.}
\label{table: dataset}
\vspace{-2ex}
\end{table}

\subsection{Baselines}

\paragraph{Fine-tuning methods.}
We categorize the fine-tuning baselines according to their characteristics as follows: \textit{(1) Topological Relationships:} \textbf{TopJudge}~\cite{DBLP:conf/emnlp/ZhongGTX0S18} explicitly models the dependency relationships among the three sub-tasks in the prediction workflow. \textit{(2) Graph-related Modeling:} \textbf{LADAN}~\cite{DBLP:conf/acl/XuWCPWZ20} designs a graph distillation module to distinguish confusing law articles. \textit{(3) Fact Decomposition: } \textbf{NeurJudge}~\cite{DBLP:conf/sigir/Yue0JWZACYW21} decomposes textual fact into different representations for each sub-task. \textit{(4) Different Pre-trained Language Models:} For the ``Text-to-Class'' style, we select \textbf{BERT}~\cite{DBLP:conf/naacl/DevlinCLT19} and \textbf{Lawformer}~\cite{DBLP:journals/aiopen/XiaoHLTS21}, while for the ``Text-to-Text'' style, we choose \textbf{mT5}~\cite{DBLP:conf/naacl/XueCRKASBR21} as the backbone model. \textit{(5) Hierarchical Reasoning: } \textbf{HRN}~\cite{DBLP:conf/emnlp/LyuH0ZGRCWR23} improves predictions by learning intermediate reasoning steps, but this also restricts its evaluation on the MultiLJP dataset with corresponding annotations.\textit{(6) LLM-based Fine-tuning:} \textbf{Vanilla-SFT} processes training data into a unified chat template for fine-tuning. \textbf{Finetune-CoT}~\cite{DBLP:conf/acl/HoSY23} initially generates Chain-of-Thought trajectories for each training data, then finetune the base model with the synthesized data. In addition to the above methods, we also consider approaches such as \textbf{CL4LJP}~\cite{DBLP:journals/tois/ZhangDZW23} and \textbf{CECP}~\cite{DBLP:conf/ijcai/ZhaoGXZC22}. However, these methods are excluded from our main experiment due to their limited adaptability to multi-label classification.

\begin{table*}[!t]
\renewcommand\arraystretch{1.05}
\centering
\scalebox{0.85}{\begin{tabular}{l|cccccc|cccccc}
\toprule
\multirow{3}{*}{Methods}
& \multicolumn{6}{c|}{CAIL2018} & \multicolumn{6}{c}{MultiLJP} \\
\cmidrule(lr){2-7}\cmidrule(lr){8-13}
& \multicolumn{2}{c}{Charge} & \multicolumn{2}{c}{Law Article} & \multicolumn{2}{c|}{Prison Term} & \multicolumn{2}{c}{Charge} & \multicolumn{2}{c}{Law Article} & \multicolumn{2}{c}{Prison Term} \\
\cmidrule(lr){2-3}\cmidrule(lr){4-5}\cmidrule(lr){6-7}\cmidrule(lr){8-9}\cmidrule(lr){10-11}\cmidrule(lr){12-13}
 & Acc. & Ma-F & Acc. & Ma-F & Acc. & Ma-F & Acc. & Ma-F & Acc. & Ma-F & Acc. & Ma-F \\
 \midrule
TopJudge & 65.5 & 74.1 & 68.2 & 74.3 & 32.1 & 32.4 & 67.6 & 55.7 & 73.9 & 54.1 & 36.1 & 33.1 \\
LADAN & 63.1 & 71.8 & 62.5 & 71.0 & 30.1 & 31.2 & 60.4 & 43.2 & 68.2 & 49.0 & 35.1 & 34.6 \\
NeurJudge & 65.7 & 71.4 & 67.4 & 70.9 & 29.6 & 33.2 & 64.8 & 51.2 & 71.8 & 55.7 & 33.9 & 32.0 \\
BERT & 64.6 & 74.6 & 68.3 & 73.5 & 31.7 & 33.5 & 66.3 & 54.2 & 73.6 & 54.0 & 35.6 & 32.9 \\
Lawformer & 66.2 & 73.1 & 67.5 & 74.4 & 30.4 & 30.7 & 68.1 & 53.8 & 76.2 & 53.8 & 36.1 & 34.7 \\
mT5 & 72.3 & 77.5 & 73.2 & 74.4 & 33.9 & 30.8 & 78.4 & 44.6 & 82.9 & 44.1 & 30.7 & 20.3 \\
HRN & - & - & - & - & - & - & 83.5 & 60.9 & 84.3 & 62.1 & 34.3 & 33.4 \\
\midrule
\multicolumn{13}{c}{\textit{LLM-based Fine-tuning}} \\
Vanilla-SFT & 74.1 & 78.6 & 74.0 & 75.5 & 32.0 & 31.3 & 85.4 & 65.2 & 87.7 & 63.5 & 32.0 & 31.3 \\
Finetune-CoT & 74.8 & 79.3 & 75.6 & 77.7 & 31.5 & 31.9 & 86.2 & 66.7 & 88.0 & 64.8 & 32.4 & 32.7 \\
\midrule
ADAPT (Ours) & \textbf{77.9} & \textbf{83.0} & \textbf{78.3} & \textbf{80.0} & \textbf{37.9} & \textbf{35.8} & \textbf{90.3} & \textbf{73.1} & \textbf{91.1} & \textbf{75.4} & \textbf{37.3} & \textbf{35.2} \\

\bottomrule
\end{tabular}}
\caption{Experimental results on the fine-tuning setting. The best results are in bold.}
\label{table:ft_main_result}
\end{table*}

\begin{table*}[!t]
\renewcommand\arraystretch{1.05}
\centering
\scalebox{0.82}{\begin{tabular}{lcc|cccc|cccc}
\toprule
\multirow{2}{*}{Model} & \multirow{2}{*}{Params.} & \multirow{2}{*}{Demos.} & \multicolumn{4}{c|}{Charge} & \multicolumn{4}{c}{Law Article} \\
\cmidrule(lr){4-7} \cmidrule(lr){8-11}
 & & & Acc. & Ma-P & Ma-R & Ma-F & Acc. & Ma-P & Ma-R & Ma-F \\
\midrule
\multicolumn{11}{c}{\textit{In-domain LLMs}} \\
Disc-LawLLM & 13B & \ding{55} & 44.2 & 59.7 & 61.8 & 56.6 & 55.0 & 54.5 & 70.5 & 57.7 \\
\midrule
\multicolumn{11}{c}{\textit{General-purpose LLMs}} \\
Qwen2-7B & 7B & \ding{55} & 41.7 & \underline{56.3} & 58.6 & \underline{53.0} & \underline{50.4} & \underline{51.9} & \underline{64.7} & \underline{52.8} \\
+ Few-shot & - & \ding{51} & 44.4 & 55.1 & 56.9 & 50.7 & 48.7 & 49.8 & 60.6 & 48.7  \\
+ CoT & - & \ding{51} & \textbf{45.7} & 54.7 & \underline{58.7} & 52.1 & 49.4 & 47.6 & 60.2 & 47.5  \\
+ ADAPT & - & \ding{51} & \underline{45.0} & \textbf{59.2} & \textbf{59.4} & \textbf{55.2} & \textbf{53.3} & \textbf{56.0} & \textbf{64.8} & \textbf{55.6}  \\
\midrule
Qwen2-72B & 72B & \ding{55} & 56.2 & 60.9 & \underline{72.4} & \underline{63.2} & 57.7 & 57.2 & \underline{71.3} & 59.4 \\
+ Few-shot & - & \ding{51} & \underline{57.4} & 61.0 & 69.8 & 61.4 & 58.6 & 54.3 & 68.5 & 57.9  \\
+ CoT & - & \ding{51} & 56.9 & \underline{61.4} & 71.4 & 62.2 & \textbf{62.9}& \underline{58.6} & \textbf{73.2} & \underline{60.3}  \\
+ ADAPT & - & \ding{51} & \textbf{58.4} & \textbf{62.3} & \textbf{73.3} & \textbf{65.0} & \underline{59.7} & \textbf{60.3} & 70.4 & \textbf{61.5}  \\
\bottomrule
\end{tabular}}
\caption{Performance on the prompting setting. The best results and the second-best results of each setting are in bold and underlined, respectively.}
\label{table: prompting results}
\end{table*}

\paragraph{Prompting setting}
We employ two types of models \textit{(1) Law Specific LLMs:} We select \textbf{Disc-LawLLM}~\cite{yue2023disclawllm} as the representative, which undergo supervised fine-tuning with high-quality task data from both legal and general scenarios. \textit{(2) General-Purpose LLMs:} We choose \textbf{Qwen2-[7B, 72B]-Instruct}~\cite{DBLP:journals/corr/abs-2309-16609} to investigate performance across different models and scales. For each model, we evaluate the performance of different prompting methods under both zero-shot and few-shot settings.

\subsection{Implementation Details}
For all LLM-based fine-tuning methods, we utilize Qwen2-7B as the foundation model. 
LoRA is adopted for parameter-efficient fine-tuning of the large language model. We apply LoRA to all linear modules of the model, with both alpha and rank set to 32. The language modeling head is also unfrozen to enhance learning. Our model is fine-tuned by 10 epochs, with a learning rate of 5e-5 and a batch size of 64. The total number of training reasoning trajectories for CAIL2018 and MultiLJP is 80,141 and 157,763, respectively. Greedy decoding is used for all generative models to enhance the stability of results.
For those generated charges that are not present in the label pool, we use BGE~\cite{arxiv23_bge} to map them to the closest charge in the pool based on their representations.

\subsection{Evaluation on the Fine-tuning Setting}

The results on the fine-tuning setting are presented in Table~\ref{table:ft_main_result}. We have the following findings:

(1) Our method outperforms all baselines across all metrics on both datasets. Overall, the LLM-based approaches show superior performance, indicating that large causal language models can adapt effectively to the tasks of legal judgment prediction with targeted fine-tuning. On the CAIL2018 dataset, our approach achieves relative accuracy and Ma-F improvements of 4.1\% and 4.7\%, respectively, over Finetune-CoT. This demonstrates that our synthetic data significantly enhances the effectiveness of legal judgment prediction.

(2) The ADAPT framework achieves a more significant advantage in charge and law article prediction. These two tasks are both multi-label classification tasks, meaning a successful shot satisfies the predicted set to be completely consistent with the label set. Our improvements suggest that discriminative reasoning offers more evidence for the precise inference of the target set. And on the other side, the performance across methods is relatively close in prison term prediction. We hypothesize that this is due to the uncertainty introduced by real-world judges' discretion, which cannot be fully mitigated by classification metrics based on rigid interval divisions.

(3) Fine-tuning with discriminative reasoning trajectories further enhances the language models' performance. Finetune-CoT similarly utilizes synthetic reasoning data to promote autoregressive learning. However, the synthesized CoT data is observed to result in sub-optimal performance. We argue that this is because the ``step-by-step reasoning'' generated by the language model does not contribute to more supporting logic. Instead, it merely discusses the plain facts in most cases. 

\begin{table*}[!t]
\renewcommand\arraystretch{1.05}
\centering
\scalebox{0.82}{\begin{tabular}{l|cccc|cccc|cccc}
\toprule
\multirow{2}{*}{Model} & \multicolumn{4}{c|}{Charge} & \multicolumn{4}{c|}{Law Article} & \multicolumn{4}{c}{Prison Term}\\
 & Acc. & Ma-P & Ma-R & Ma-F & Acc. & Ma-P & Ma-R & Ma-F & Acc. & Ma-P & Ma-R & Ma-F \\
 \midrule
\textit{w/o Ask}  & 76.3 & 81.4 & 84.5 & 81.5 & 76.4 & 78.5 & 81.4 & 79.2 & 35.7 & 33.6 & 35.3 & 33.8 \\
\textit{w/o Disc}  & 76.8 & 81.2 & 85.0 & 81.8 & 76.6 & 78.2 & 82.1 & 79.0 & 35.3 & 33.0 & 35.7 & 33.5 \\
\textit{w/o Article} & 75.7 & 80.3 & 81.9 & 80.4 & 76.0 & 78.1 & 80.7 & 78.5 & 34.9 & 32.6 & 33.9 & 32.4 \\
\textit{w/o Sentencing} & 77.6 & 82.0 & 86.4 & 82.7 & 78.0 & 79.6 & 83.0 & 79.7 & 36.0 & 34.0 & 35.6 & 34.1 \\
\textit{ADAPT$\rightarrow$Refine} & 70.2 & 74.4 & 75.3 & 74.8 & 69.5 & 72.1 & 77.1 & 74.7 & 28.9 & 29.4 & 30.7 & 29.8 \\
ADAPT (Ours) & \textbf{77.9} & \textbf{82.3} & \textbf{86.9} & \textbf{83.0} & \textbf{78.3} & \textbf{79.6} & \textbf{83.5} & \textbf{80.0} & \textbf{37.9} & \textbf{35.6} & \textbf{37.2} & \textbf{35.8} \\
\bottomrule
\end{tabular}}
\caption{Ablation results on CAIL2018. The best results are in bold.}
\label{table: ablation study}
\vspace{-2ex}
\end{table*}

\begin{figure*}[!t]
	\centering
	\includegraphics[width=0.95\linewidth]{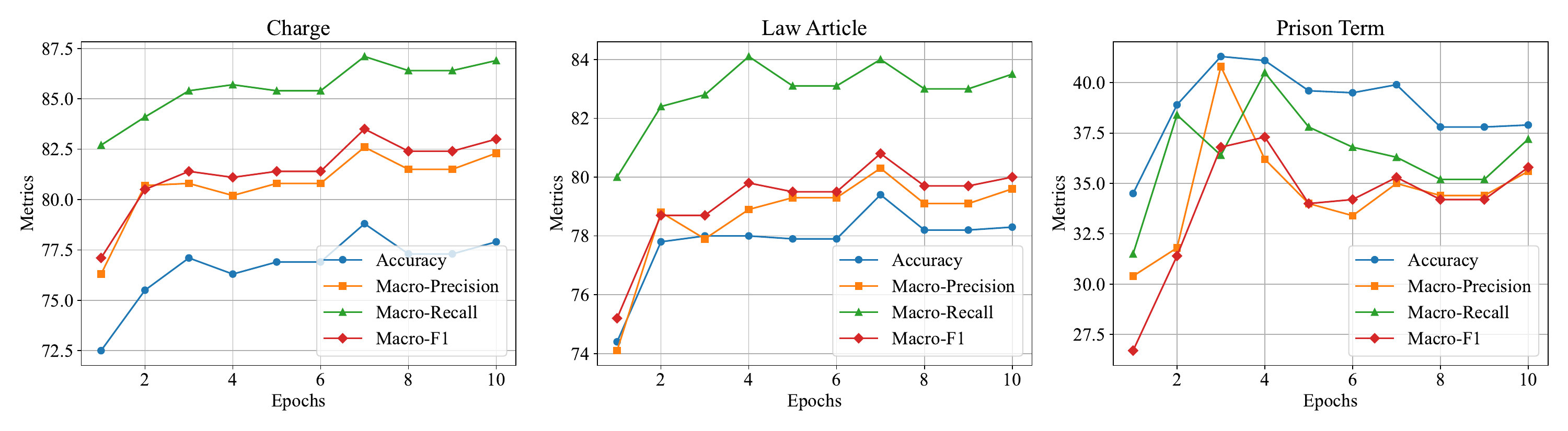}
 \caption{Fine-tuning performance of each sub-task with epochs on the CAIL2018 dataset.}
  \vspace{-2ex}
 \label{fig: epochs}
\end{figure*}

\subsection{Evaluation on the Prompting Setting}
We evaluate under the prompting setting using the CAIL2018 dataset, as the MultiLJP dataset contains only 23 charges and can not reflect the effect of confusing charges in real-world scenarios. During our experiments, we observed that the in-domain model fails to effectively adhere to instructions for few-shot learning, so we only tested its zero-shot ability. Moreover, we discard the prison term prediction task because LMs typically refuse to predict the accurate terms. The results are reported in Table~\ref{table: prompting results}, from which we can observe:

(1) Our approach generally demonstrates superior performance across models of different sizes, suggesting that prompts can still activate discriminative reasoning to some extent. Moreover, we also observe that the improvements are not pronounced as in the fine-tuning setting. This indicates that fine-tuning can further enhance the model's capabilities within our framework.

(2) Larger models typically bring better performance, but they still exhibit a notable disparity when compared to models trained for LJP specifically. This might be because recent leading open-source models have been trained on extensive data from major domains. This suggests that fine-tuning remains a valuable strategy for LLMs to adapt to the requirements of LJP tasks

\subsection{Ablation Study}
We investigate the effects of various reasoning data on the final LJP tasks within the fine-tuning context. The specific ablation strategies are described as follows: (1) \textit{w/o Ask:} The task of summarizing legal elements from facts is removed, and $y^{\text{Ask}}$ is excluded from $y^{\text{All}}$. (2) \textit{w/o Disc:} After the \textit{Ask} step, the language model must directly predict the charges and law articles. (3) \textit{w/o Article:} We remove the law article-related reasoning trajectories in the training data. (4) \textit{w/o Sentencing:} The language model no longer analyzes sentencing factors before determining the prison term prediction. (5) \textit{ADAPT$\rightarrow$Refine:} We construct candidate items for charges and law articles and provide them to the large language model, requiring it to refine them and determine the final prediction. During inference, the candidates are derived from the top-k items in the probability distribution of BERT, while during training, the correct labels are ensured to be included among these candidates.

The ablation results are shown in Table~\ref{table: ablation study}. Firstly, we can observe the removal of each sub-task leads to a performance decline, indicating that each type of synthetic data makes a positive contribution. Additionally, we notice that the ablation associated with law articles causes obvious impact. This impact may be attributed to the better alignment between legal provisions and facts of real cases, which enhances legal reasoning abilities in other forms. Finally, it is observed that the ablation related to sentencing almost exclusively affects prison term prediction. This can likely be attributed to the fact that sentencing factors and conviction factors are orthogonal in most cases.

Moreover, we find that \textit{ADAPT$\rightarrow$Refine} results in a significant performance decline. This is because the top-k candidates provided by the domain model (i.e., BERT) during inference often do not include the ground truth labels, thus the LLM can only select relatively close labels in such cases. We think that the fine-tuned LLMs are inherently capable of generating high-quality candidates and don't require assistance from smaller external models.

\begin{figure*}[!t]
	\centering
	\includegraphics[width=0.95\linewidth]{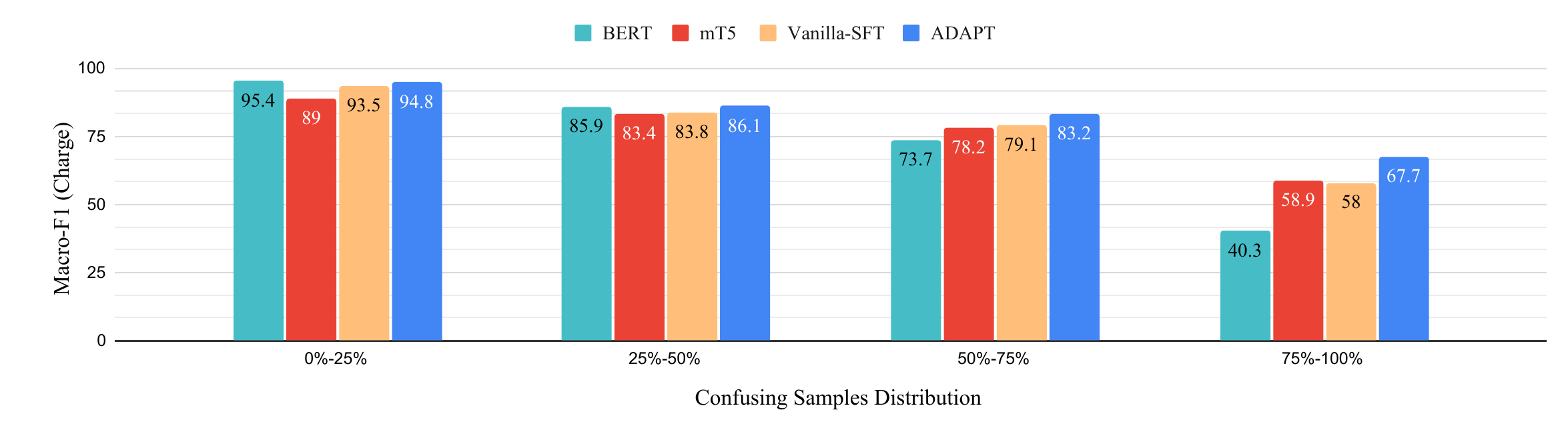}
	\caption{Performance in different charge subgroups of various methods. Intervals of larger numbers (e.g., 75\%-100\%) indicate greater difficulty of the corresponding subgroup.}
	\label{fig: confusing experiments}
 \vspace{-2ex}
\end{figure*}

\subsection{Effect of Training Epochs}
Instruction tuning has been observed to lead to performance degradation with an increasing number of training epochs. However, previous studies demonstrate that it often requires many epochs (e.g., 20 epochs) for model performance to coverage in LJP tasks.
We investigate the effects of training epochs for the ADAPT framework in this section. The results, presented in Figure~\ref{fig: epochs}, reveal that:

(1) In the predictions of charges and law articles, metrics gradually increase with the number of training epochs and stabilize after reaching a peak at a certain epoch. This indicates that the model can continuously learn effective features from the data even after the initial complete iteration.

(2) The metric curves for different sub-tasks exhibit notable distinctions. Unlike the other two tasks, most of the metrics for prison term prediction show a marked decline after the third epoch. This divergence highlights the inherent difficulty in predicting prison terms. Judicial discretion in sentencing can span multiple pre-defined intervals, thereby increasing the risk of the model overfitting when learning from repeated samples.

(3) In multi-label classification tasks (i.e., charge and law article prediction), the model consistently exhibits higher performance in the Macro-Recall compared to the other three metrics. This indicates that the fine-tuned language model tends to identify more possible positive candidates. We believe this also suggests the potential for further refining results in our proposed reasoning framework.

\subsection{Performance on Different Difficulty}
Exploring performance across different charges can help us better understand the detailed improvements achieved by our method. We 
 first calculate the F1 scores of the finetuned BERT model for each charge in the CAIL2018 dataset and rank these charges from highest to lowest. For clearer visualization, we categorize the ranked charges into four sets and then evaluate the macro-F1 scores of all finetuned models on the charge prediction task. Generally, the charges in the higher quartiles (e.g., 75\%-100\%) exhibit greater prediction difficulty, as evidenced by the poor performance of mainstream "Text-to-Label" style models on these charges. The experimental results are shown in Figure~\ref{fig: confusing experiments}, from which we can observe the following findings:

(1) Our ADAPT framework achieves a more significant improvement on difficult sets. Specifically, in the 75\%-100\% interval, ADAPT achieves relative improvements of 15.1\% and 16.7\% compared to mT5 and Vanilla-SFT, respectively. Conversely, in the 0\%-25\% interval, the relative improvements are merely 6.5\% and 1.4\%. This suggests that our discriminative reasoning approach effectively delineates subtle differences between various charges, thereby significantly enhancing prediction accuracy for more confusing charges.

(2) The marginal benefits of employing simple instruction fine-tuning on larger models are limited. We can observe that despite Vanilla-SFT leveraging a language model with 7B parameters, its improvements over mT5 are not substantial. Notably, in cases involving more difficult charges, Vanilla-SFT is even likely to demonstrate a slight decline in performance. This finding highlights the importance of identifying an appropriate reasoning pattern to enhance the effectiveness of large language models in legal judgment prediction.


\section{Conclusion}
In this paper, we presented a novel framework to enable discriminative reasoning in LLMs for legal judgment prediction. Our ADAPT framework effectively distinguishes confusing charges by determining the degree of alignment between each candidate charge and the criminal facts before the final prediction. Furthermore, we utilize multi-task instruction tuning on synthetic data to enhance the language model's comprehension of this reasoning pattern. Extensive experiments demonstrate the effectiveness of our approach, particularly its robustness in handling confusing charges. We believe that our work will improve the integration of LLMs in legal judgment prediction and contribute to the community's understanding of reasoning patterns in specific tasks.

\section{Limitations}
Despite the promising results that have been demonstrated in our framework, several limitations must be acknowledged:

\paragraph{Cost of Synthesized Data.}
Our method requires the synthesis of reasoning trajectories from existing judgment data to facilitate fine-tuning, potentially leading to increased computational expenses or API costs. Fortunately, the cost of utilizing large language models is rapidly decreasing. For instance, recent inference services such as Qwen and DeepSeek require less than \$0.0001 per 1,000 tokens. Therefore, we believe that generating reasoning data at this scale is entirely acceptable.

\paragraph{Limited Scope of Open Datasets.}
Our framework demonstrates strong generalization capabilities for the case types it was trained on. However, the most diverse dataset we employed, CAIL2018, encompasses fewer than 200 distinct charges, potentially limiting its applicability in all real-world scenarios. Consequently, we recommend further synthesizing more comprehensive reasoning trajectories using public or private domain data to meet the specific needs of real-world applications.

\paragraph{Potential Dataset Leakage Risks.}
Although the large language models utilized in our experiments are open-source, the datasets employed during their training are not entirely transparent, potentially posing risks of data leakage. Our solution is to evaluate various methods on the same foundation model to ensure fair comparisons. The relative improvements in various settings can prove our advantage.

\section{Ethical Discussion}
\paragraph{Privacy and Data Security.}
Legal data often includes sensitive and confidential information about individuals and entities. Mishandling this data can lead to serious privacy breaches. The two datasets adopted in our experiments are robustly anonymized to protect this information. 

\paragraph{Potential Bias in Training Data.}
Large language models may learn bias from judgments in the training set. In a real-world case, the final judgment can be affected by some factors like social comments, the judge's style, etc. We need to test and identify possible biases before application. 

\paragraph{Legal and Ethical Compliance.}
Adhering to existing legal and ethical standards is essential when deploying LLMs for legal judgment prediction. We advise users to critically evaluate the model's predictions and make independent decisions about their adoption, rather than uncritically accepting the machine's reasoning.


\appendix
\section*{Appendix}

\section{Prompts}
\label{Appendix: Prompts}
In this section, we provide prompts for multi-task instruction tuning and synthesizing data, respectively. The text in pink denotes the input information.

\begin{figure}[!t]
	\centering
	\includegraphics[width=0.95\linewidth]{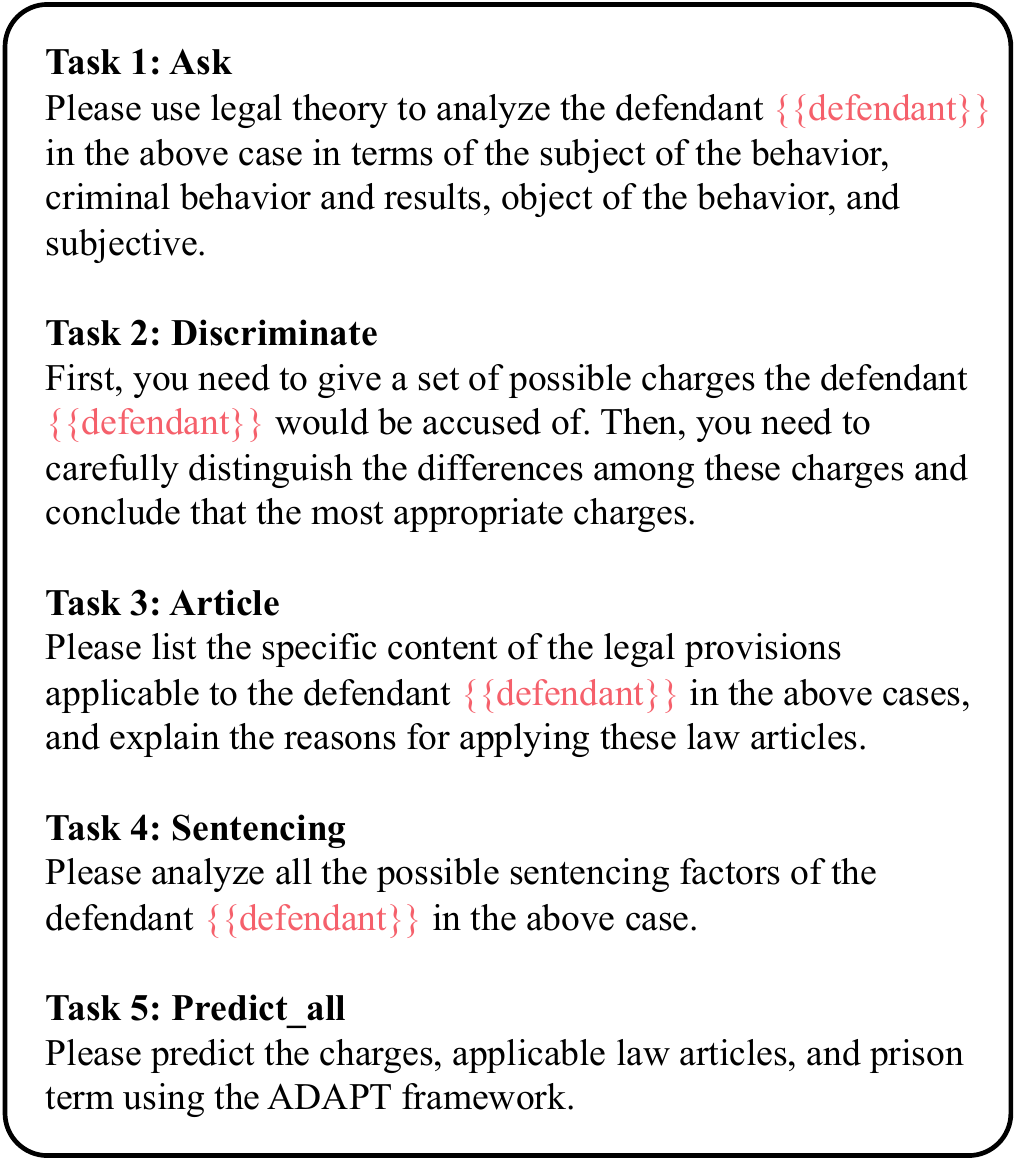}
	\caption{Prompt for each task of our multi-task instruction tuning.}
\end{figure}

\begin{figure}[!t]
	\centering
	\includegraphics[width=0.95\linewidth]{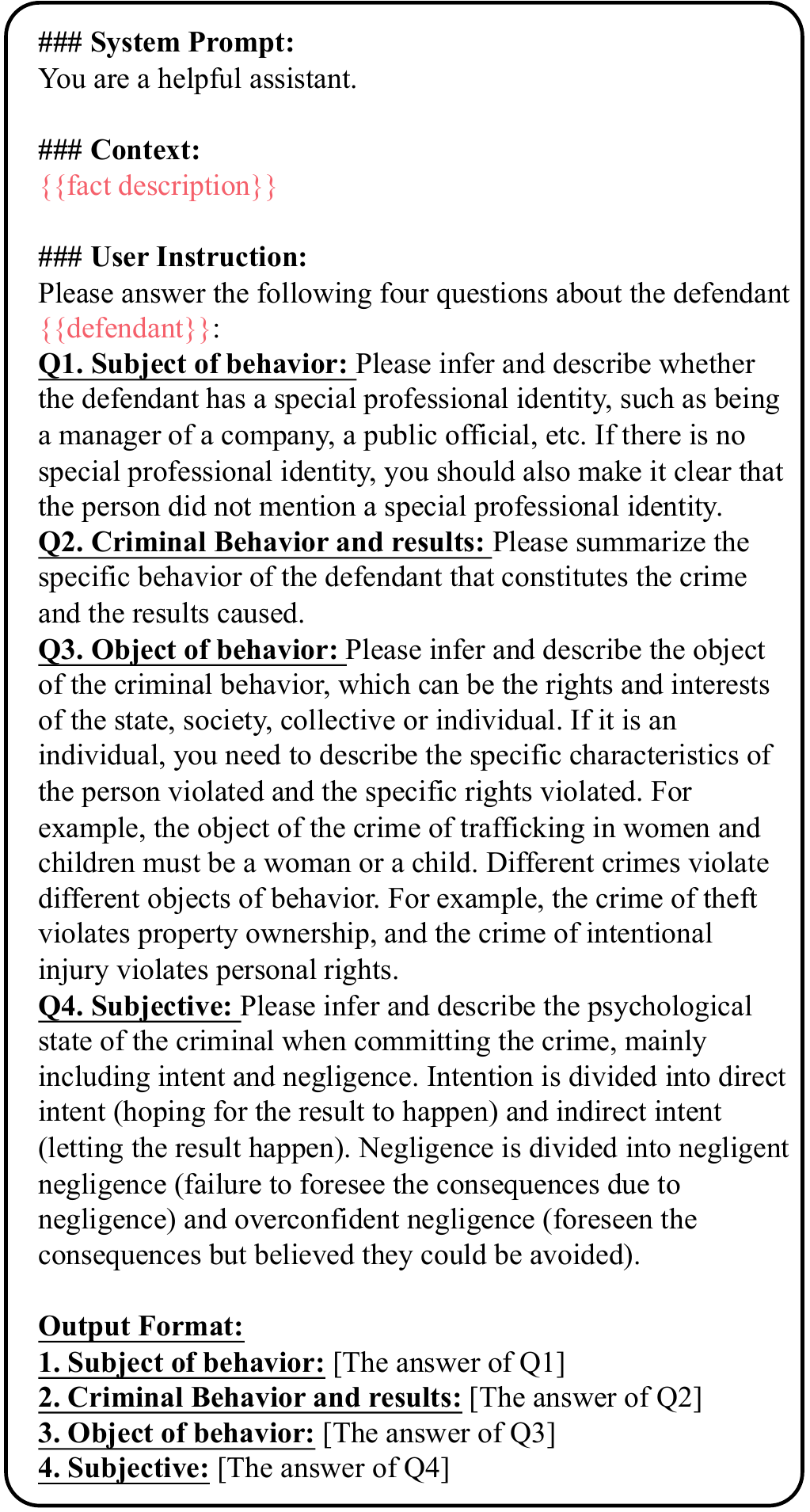}
	\caption{Prompt for synthesizing the trajectory of the step \textit{Ask}.}
\end{figure}

\begin{figure}[!t]
	\centering
	\includegraphics[width=0.95\linewidth]{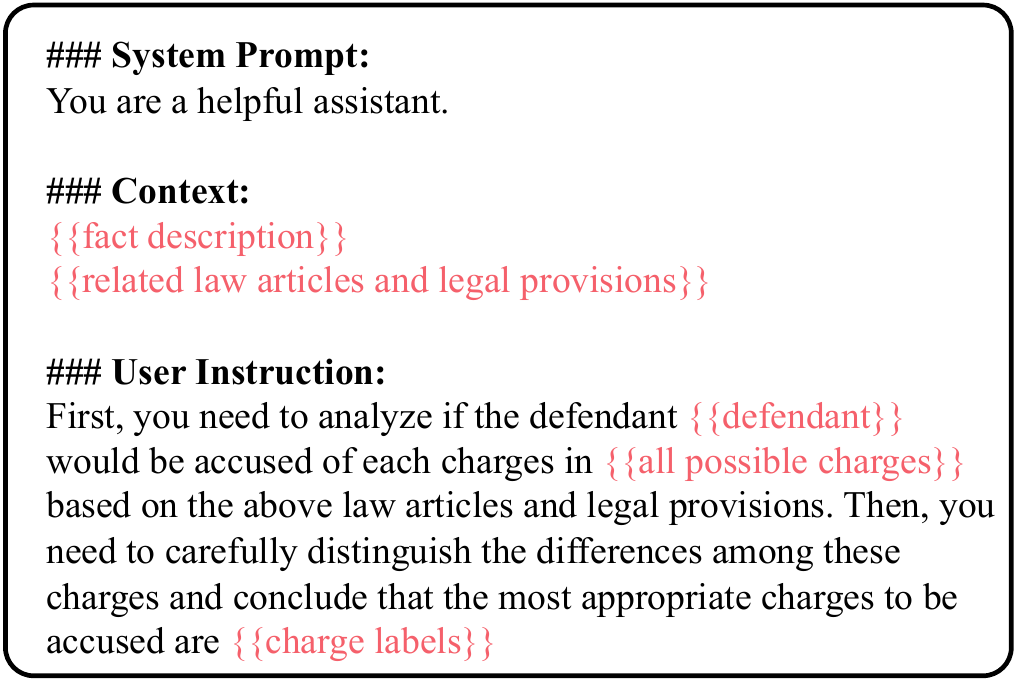}
	\caption{Prompt for synthesizing the trajectory of the step \textit{Discriminate}.}
\end{figure}

\begin{figure}[!t]
	\centering
	\includegraphics[width=0.95\linewidth]{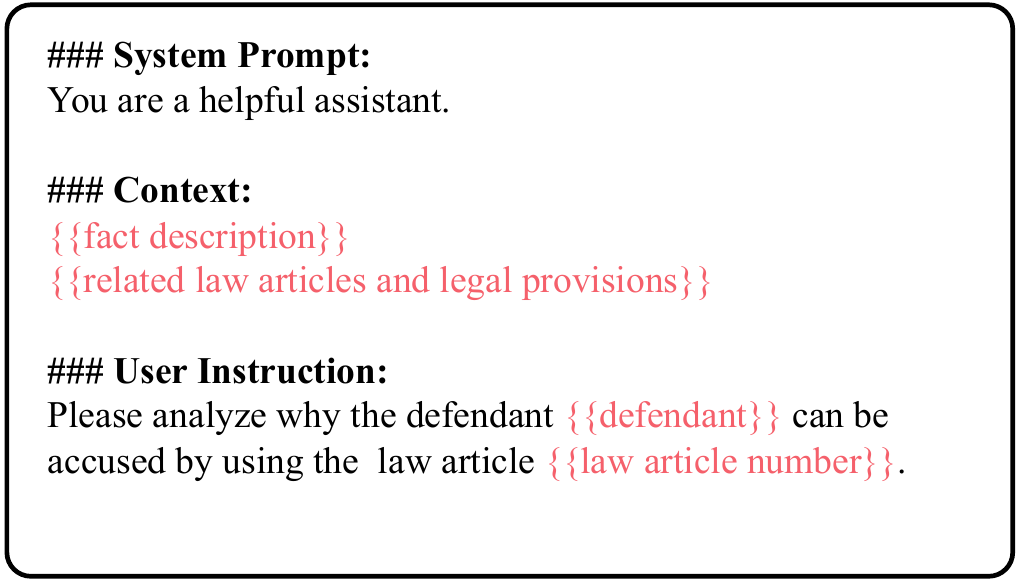}
	\caption{Prompt for synthesizing the trajectory of the task \textit{Article}.}
\end{figure}

\begin{figure}[!t]
	\centering
	\includegraphics[width=0.95\linewidth]{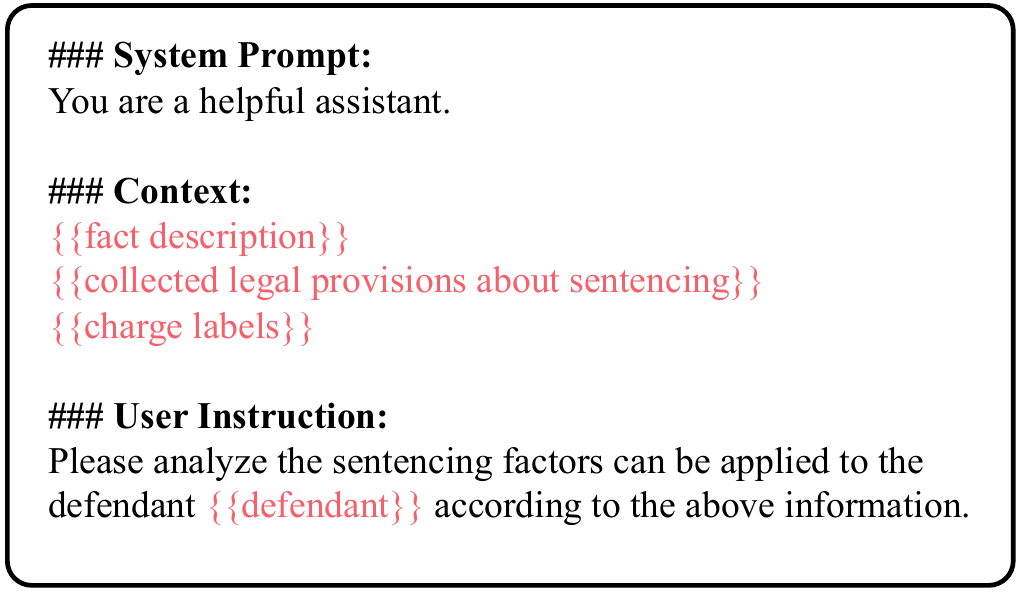}
	\caption{Prompt for synthesizing the trajectory of the task \textit{Sentencing}.}
\end{figure}

\end{document}